\documentclass[letterpaper, 10 pt, conference]{ieeeconf}
\IEEEoverridecommandlockouts 
\usepackage{amsmath} 
\usepackage{amssymb}
\usepackage[mathscr]{euscript}
\usepackage{graphicx}
\usepackage{xcolor}
\usepackage{algorithm}
\usepackage[noend]{algpseudocode}
\usepackage{tabularx}
\usepackage{multirow}
\usepackage{hhline}
\usepackage{geometry}

\usepackage{tikz}
\usepackage{textcomp}
\usepackage{hyperref}
\usepackage{lipsum}
\newcommand\copyrighttext{%
  \footnotesize \textcopyright 2020 IEEE. Personal use of this material is permitted.
  Permission from IEEE must be obtained for all other uses, in any current or future
  media, including reprinting/republishing this material for advertising or promotional
  purposes, creating new collective works, for resale or redistribution to servers or
  lists, or reuse of any copyrighted component of this work in other works.}
\newcommand\copyrightnotice{%
\begin{tikzpicture}[remember picture,overlay]
\node[anchor=south,yshift=10pt] at (current page.south) {\fbox{\parbox{\dimexpr\textwidth-\fboxsep-\fboxrule\relax}{\copyrighttext}}};
\end{tikzpicture}%
}

\providecommand{\image}[3]{
	\begin{figure}
      \centering
      \vspace*{6pt}
      \includegraphics[width=#2mm]{./images/#1}
      \caption{#3} \label{fig:#1}
	\end{figure}
}

\usepackage[capitalise,nameinlink]{cleveref}
  \crefname{section}{Sect.}{Sect.}
  \Crefname{section}{Section}{Sections}
  \crefname{figure}{Fig.}{Fig.}
  \Crefname{figure}{Figure}{Figures}
  \crefname{table}{Tabl.}{Tabl.}
  \Crefname{table}{Table}{Tables}
\newcommand{\squeezeup}{\vspace{-2.5mm}}
\title{Learning Force Control for Contact-rich Manipulation Tasks \\with Rigid Position-controlled Robots}

\begin{document}
\makeatletter
\newcommand{\linebreakand}{%
  \end{@IEEEauthorhalign}
  \hfill\mbox{}\par
  \mbox{}\hfill\begin{@IEEEauthorhalign}
}
\makeatother

\author{
 Cristian C. Beltran-Hernandez$^{{1}{*}}$ \quad
 Damien Petit$^{1}$ \quad
 Ixchel G. Ramirez-Alpizar$^{2,1}$ \quad \\
 Takayuki Nishi$^{3}$ \quad Shinichi Kikuchi$^{3}$  \quad
 Takamitsu Matsubara $^{4}$ \quad
 Kensuke Harada$^{1,2}$ %
\thanks{$^{1}$ 
Graduate School of Engineering Science, Osaka University, Japan.}
\thanks{$^{2}$ Automation Research Team, Industrial CPS Research Center, National Institute of Advanced Industrial Science and Technology (AIST), Japan.}
\thanks{$^{3}$ Process Engineering \& Technology Center, Research \& Development Management Headquarters, FUJIFILM Corporation, Japan}
\thanks{$^{4}$ Robot Learning Laboratory, Institute for Research Initiatives, Nara Institute of Science and Technology (NAIST), Japan}
\thanks{$^{*}$ Corresponding author e-mail: \newline \hspace*{10mm} {\tt beltran[at]hlab.sys.es.osaka-u.ac.jp}}}

\maketitle
\copyrightnotice
\setlength{\textfloatsep}{0.5\baselineskip plus 0.2\baselineskip minus 0.5\baselineskip}

\begin{abstract}
Reinforcement Learning (RL) methods have been proven successful in solving manipulation tasks autonomously. However, RL is still not widely adopted on real robotic systems because working with real hardware entails additional challenges, especially when using rigid position-controlled manipulators. These challenges include the need for a robust controller to avoid undesired behavior, that risk damaging the robot and its environment, and constant supervision from a human operator. 
The main contributions of this work are, first, we proposed a learning-based force control framework combining RL techniques with traditional force control. Within said control scheme, we implemented two different conventional approaches to achieve force control with position-controlled robots; one is a modified parallel position/force control, and the other is an admittance control. Secondly, we empirically study both control schemes when used as the action space of the RL agent. Thirdly, we developed a fail-safe mechanism for safely training an RL agent on manipulation tasks using a real rigid robot manipulator. The proposed methods are validated on simulation and a real robot, an UR3 e-series robotic arm.
\end{abstract}

\section{Introduction}
 In the age of the 4th industrial revolution, there is much interest in applying artificial intelligence to automate industrial manufacturing processes. Robotics, in particular, holds the promise of helping to automate processes by performing complex manipulation tasks. Nevertheless, safely solving complex manipulation tasks in an unstructured environment using robots is still an open problem\cite{kroemer2019review}.

 Reinforcement learning (RL) methods have been proven successful in solving manipulation tasks by learning complex behaviors autonomously in a variety of tasks such as grasping \cite{pmlr-v87-kalashnikov18a,levine2018hand-eye}, pick-and-place \cite{gu2017deep}, and assembly \cite{thomas2018learning}. While there are some instances of RL research validated on real robotic systems, most works are still confined to simulated environments due to the additional challenges presented by working on real hardware, especially when using rigid position-controlled robots. These challenges include the need for a robust controller to avoid undesired behavior that risk collision with the environment, and constant supervision from a human operator.
 
 So far, when using real robotic systems with RL, there are two common approaches. The first approach consists of learning high-level control policies of the manipulator. Said approach assumes the existence of a low-level controller that can solve the RL agent's high-level commands. Some examples include agents that learn to grasp \cite{pmlr-v87-kalashnikov18a,levine2018hand-eye} or to throw objects \cite{zeng2019tossingbot}. In said cases, the agent learns high-level policies, e.g., learns the position of the target object and the grasping pose, while a low-level controller, such as a motion planner, directly controls the manipulator's joints or end-effector position. Nevertheless, the low-level controller is not always available or easy to manually engineer for each task, especially for achieving contact-rich manipulation tasks with a position-controlled robot.  The second approach is to learn low-level control policies using soft robots \cite{levine2016end,schoettler2019deep,johannsmeier2019framework}, manipulators with joint torque control or flexible joints, which are considerably safer to work with due to their compliant nature, particularly in the case of allowing an RL agent to explore its surroundings where collisions with the environment may be unavoidable. Our main concern with this approach is that most industrial robot manipulators are, by contrast, rigid robots (position-controlled manipulators). Rigid robots usually run on position control, which works well for contact-free tasks, such as robotic welding, or spray-painting \cite{lynch2017modern}. However, they are inherently unsuitable for contact-rich manipulation tasks since any contact with the environment would be considered as a disturbance by the controller, which would generate a collision with a large contact force. Force control methods \cite{siciliano2012robot} can be used to enable the rigid manipulator to perform tasks that require contact with the environment, though the controller's parameters need to be properly tuned, which is still a challenging task.
 Therefore, we propose a method to safely learn low-level force control policies with RL on a position-controlled robot manipulator. 
 
 This paper presents three main contributions. First, a control framework for learning low-level force control policies combining RL techniques with traditional force control. Within said control scheme, we implemented two different conventional force control approaches with position-controlled robots; one is a modified parallel position/force control, and the other is an admittance control. Secondly, we empirically study both control schemes when used as the action space of the RL agent. Thirdly, we developed a fail-safe mechanism for safely training an RL agent on manipulation tasks using a real rigid robot manipulator. The proposed methods are validated on simulation and real hardware using a UR3 e-series robotic arm.

\section{Related Work}\label{sec:related-work}
\subsubsection{Force control}
Force control methods address the problem of interaction between a robot manipulator and its environment, even in the presence of some uncertainty (geometric and dynamic constraints) on contact-rich tasks \cite{chiaverini1993parallel, impedance1984}. These methods provide direct control of the interaction through contact force feedback and a set of parameters, which describe the dynamic interaction between the manipulator and the environment. However, prior knowledge of the environment is necessary to properly define the controller's parameters at each phase of the task, such as stiffness. Existing methods address said problem by either scheduling variable gains\cite{mitrovic2011learning}, using adaptive methods for setting the gains \cite{chien2004adaptive}, or learning the gains from demonstrations \cite{racca2016learning}. Instead, we propose to directly learn the time-variant force control gains from interacting with the environment.

\subsubsection{Reinforcement learning and force control}
Previous research has also studied the use of RL methods to learn force control gains. Buchli et al.  \cite{buchli2011learning}  uses policy improvements with path integrals (PI2) \cite{theodorou2010reinforcement} to refine initial motion trajectories and learn variable scheduling for the joint impedance parameters. 
Similarly, Bogdanovic et al. \cite{bogdanovic2019learning}, proposed a variable impedance control in joint-space, where the gains are learned with Deep Deterministic Policy Gradient (DDPG) \cite{lillicrap2015continuous}. Likewise, Mart\'in-Mart\'in et al \cite{martin2019iros}, proposed a variable impedance control in end-effector space (VICES). 
 
However, in all these cases, control of the manipulator's joint torques is assumed, which is not available for most industrial manipulators. Instead, we focus on position-controlled robot manipulators and provide a method to learn manipulation tasks using force feedback control where the controller gains are learned through RL methods. 
Luo et al. \cite{luo2018deep} propose a method for achieving peg-in-hole tasks on a deformable surface using RL and validated their approach on a position-controlled robot. They propose learning the motion trajectory based on the contact force information. However, the tuning of the compliant controller's parameters is not taken into account. We are proposing a method for learning not only the motion trajectory based on force feedback but simultaneously fine-tuning the compliant controller's parameters.

Additionally, both Bogdanovic \cite{bogdanovic2019learning} and Mart\'in-Mart\'in \cite{martin2019iros} study the importance of different action representation in RL for contact-rich robot manipulation tasks. We similarly provide an empirical study comparing different choices of action space based on force feedback control methods for rigid robots on contact-rich manipulation tasks.

\subsubsection{Learning with real-world manipulators}
Some research projects have explored the capabilities of RL methods on real robots by testing them on a large scale, such as Levine et al. \cite{levine2018hand-eye} and Pinto et al. \cite{pinto2016supersizing}, both in which a massive amount of data was collected for learning robotic grasping tasks. However, in both works, a high-level objective, grasp posture, is learned from the experience obtained. In contrast, contact-rich tasks require learning direct low-level control to, for example, reduce contact force for safety reasons. 
On the other hand, Mahmood et al. \cite{mahmood18a} propose a benchmark for learning policies on real-world robots, so different RL algorithms can be evaluated on a variety of tasks. Nevertheless, the tasks available in \cite{mahmood18a} are either locomotion tasks with a mobile robot or contact-free tasks with a robot manipulator. In this work, we propose a framework for learning contact-rich manipulation tasks with real-world robot manipulators based on force control methods.
\vspace{-2mm}
\section{Methodology}\label{sec:method}
    The present study deals with high precision assembly tasks with a position-controlled industrial robot. Due to the difficulty of obtaining a precise model of the physical interaction between the robot and its environment, RL is used to learn both the motion trajectory and the optimal parameters of a compliant controller. The RL problem is described in \Cref{sec:reinforcement-learning}. The architecture of the system and the interaction control methods considered are explained in \Cref{subsec:pose-representation}, \Cref{subsec:learning-interaction-control}, and \Cref{subsec:interaction-ctrl-implementation}. Finally, our safety mechanism that allows the robot to learn unsupervised is described in \Cref{subsec:safe-framework}.

\subsection{Reinforcement Learning}\label{sec:reinforcement-learning}
    Robotic reinforcement learning is a control problem where a robot, the agent, acts in a stochastic environment by sequentially choosing actions over a sequence of time steps. The goal is to maximize  a cumulative reward. Said problem is modeled as a Markov Decision Process. The environment is described by a state $\textbf{s} \in \mathscr{S}$. The agent can perform actions $\textbf{a} \in A$, and perceives the environment through observations $\textbf{o} \in O$, which may or not be equal to $\textbf{s}$. We consider an episodic interaction of finite time steps with a limit of $T$ time steps per episode. The agent's goal is to find a policy $\pi(\textbf{a}(t) \,|\, \textbf{o}(t))$ that selects actions $\textbf{a}(t)$ conditioned on the observations $\textbf{o}(t)$ to control the dynamical system. Given an stochastic dynamics $p(\textbf{s}(t+1) \,|\, \textbf{s}(t), \textbf{a}(t))$ and a reward function $r(\textbf{s}, \textbf{a})$, the aim is to find a policy $\pi*$ that maximizes the expected sum of future rewards given by $R(t)=\sum_i^{\infty} \gamma r(s(t),a(t))$ with $\gamma$ being a discount factor \cite{sutton2018reinforcement}.

\subsubsection*{Soft-Actor-Critic}
    We use the state-of-the-art model-free RL method called Soft-Actor-Critic (SAC) \cite{Haarnoja2018SoftAO}. SAC is an off-policy actor-critic deep RL algorithm based on the maximum entropy reinforcement learning framework. SAC aims to maximize the expected reward while optimizing a maximum entropy. The SAC agent optimizes a maximum entropy objective, which encourages exploration according to a temperature parameter $\alpha$. The core idea of this method is to succeed at the task while acting as randomly as possible. Since SAC is an off-policy algorithm, it can use a replay buffer to reuse information from recent rollouts for sample-efficient training. We use the SAC implementation from TF2RL\footnote{TF2RL: RL library using TensorFlow 2.0. https://github.com/keiohta/tf2rl}.

\subsection{System overview}\label{subsec:system-overview}
\image{}{80}{
Proposed learning force control scheme. The input to the system is a goal end-effector pose, $\textbf{x}_g$. The policy actions are trajectory commands, $\textbf{a}_x$, and parameters, $\textbf{a}_p$, of a force controller. 
}

    Our proposed method aims to combine a force control with RL to learn contact-rich tasks when using position-controlled robots. \Cref{fig:control-scheme} describes the proposed control scheme combining an RL policy and a force control method. We assume knowledge of the goal pose of the robot's end-effector, $\textbf{x}_g$. Both the policy and the force controller receive as feedback the pose error, $\textbf{x}_e=\textbf{x}_g-\textbf{x}$, and the contact force $F_{ext}$. The velocity of the end-effector, $\Dot{\textbf{x}}$, is also included in the policy's observations. The F/T sensor signal is filtered using a simple low-pass filter.
    
    The force control method has two internal controllers. First, a PD controller that generates part of the motion trajectory based on the pose error, $\textbf{x}_e$. Second, a force feedback controller that alters the motion trajectory according to the perceived contact force, $F_{ext}$.
    
    The RL policy has two objectives. First, to generate a motion trajectory, $\textbf{a}_x$. \Cref{fig:problem-statement}, shows how a simple P-controller (from the force control method)  would not be enough to solve the task without producing a collision with the environment. For most cases, the P-controller trajectory would just attempt to penetrate the environment, since knowledge of the environment's geometry is not assumed. Nevertheless, the P-controller trajectory is good enough to speed up the agent's learning since it is already driven towards the goal pose. Therefore, to achieve the desired behavior, the nominal trajectory of the robot is the combination of the P-controller trajectory with the policy's trajectory. The second objective of the policy is to fine-tune the force control methods parameters, $\textbf{a}_p$, to minimize the contact force when it occurs. We defined a collision as exceeding a maximum contact force in any direction. Therefore, contact with the environment is acceptable, but the policy's second goal is to avoid collisions. The policy also controls the P-controller's gains; thus, the policy decides how much to rely on the P-controller trajectory.
    
\subsubsection{Pose Control Representation}\label{subsec:pose-representation}

    The pose of the robot's end-effector is given by $\textbf{x} = [\textbf{p},\phi]$, where $\textbf{p} \in \mathbb{R}^3 $ is the position vector and $\phi \in \mathbb{R}^4$ is the orientation vector. The orientation vector is described using Euler parameters (unit quaternions) denoted as $\phi = \{\eta, \varepsilon\}$; where $\eta \in \mathbb{R}$ is the scalar part of the quaternion and $\varepsilon \in \mathbb{R}^3$ the vector part. Using unit quaternions allows the definition of a proper orientation error for control purposes with a fast computation compared to  using rotation matrices \cite{campa2008unit}.
    
    The position command from the force controller is $\textbf{x}_c = [p_t,\phi_t]$, where $p_t$ is the commanded translation, and $\phi_t$ is the commanded orientation for the time step $t$. The desired joint configuration for the current time step, $\textbf{q}_c$, is obtain from an Inverse Kinematics (IK) solver based on $\textbf{x}_c$.

\subsubsection{Learning force control}\label{subsec:learning-interaction-control}

    Two of the most common force control schemes are considered in these work, parallel position/force control \cite{chiaverini1993parallel} and admittance control \cite{impedance1984}. The main drawback of said control schemes is the requirement to tune the parameters for each specific task properly. Changes in the environment (e.g., surface stiffness) may require a new set of parameters. Thus, we propose a self-tuning process using RL method.

    The policy actions are $\textbf{a} = [\textbf{a}_x, \textbf{a}_p]$,
    where $\textbf{a}_x = [\textbf{p}, \phi]$ are position/orientation commands, and $\textbf{a}_p$ are controller's parameters. $\textbf{a}_p$ is different and specific for each type of controller, see \Cref{subsec:parallel-scheme} and \Cref{subsec:admittance-scheme} for details. The policy has a control frequency of 20 Hz while the force controller has a control frequency of 500 Hz.

\image{}{87}{Proposed approach to solve contact-rich tasks. Assuming knowledge of the goal pose of the robot's end-effector, a simple P-controller can be designed. Our approach aims to combine this knowledge with the policy to generate the motion trajectory. }

\subsection{Force control implementation} \label{subsec:interaction-ctrl-implementation}

\subsubsection{PID parallel Position/Force Control}\label{subsec:parallel-scheme}

    Based on \cite{chiaverini1993parallel}, we implemented a PID parallel position/force control with the addition of a selection matrix to define the degree of control of position and force over each direction, as shown in \Cref{fig:parallel-scheme}. The control law consists of a PD action on position, a PI action on force, a selection matrix and the policy position action, $\textbf{a}_x$,
    \begin{equation}
    \begin{aligned}
        u ~=  S (K_p^x\textbf{x}_e &+ K_d^x\Dot{\textbf{x}_e}) + \textbf{a}_x + \\
         & (I-S)(K_p^fF_{ext}+K_i^f\int F_{ext} dt)
    \end{aligned}
    \end{equation}
    where $u$ is the vector of driving generalized forces. The selection matrix is \[S = diag(s_1, ..., s_6),\quad s_j \in [0,1]\] 
    where the values correspond to the degree of control that each controller has over a given direction.

    Our parallel control scheme has a total of 30 parameters, 12 from the position PD controller's gains, 12 from the force PI controller's (PI) gains, and 6 from the selection matrix $S$. We reduced the number of controllable parameters to prevent unstable behavior and to reduce the system's complexity. For the PD controller, only the proportional gain, $K_p^x$, is controllable while the derivative gain, $K_d^x$, is computed based on the $K_p^x$. $K_d^x$ is set to have a critically damped relationship as
    \[ K_d^x = 2\sqrt{K_p^x} \]
    Similarly, for the PI controller, only the proportional gain, $K_p^f$, is controllable, the integral gain $K_i^f$ is computed with respect to $K_p^f$. In our experiments, $K_i^f$ was set empirically to be $1\%$ of $K_p^f$. In total, 18 parameters are controllable. 

    In summary, the policy actions regarding the parallel controller's parameters are $\textbf{a}_p = [K_p^x, K_p^f, S]$.
    
    To narrow the agents choices for the force control parameters, we follow a similar strategy as in \cite{bogdanovic2019learning}. Assuming we have access to some baseline gain values, $P_{\text{base}}$. We then define a range of potential values for each parameter as $[P_{\text{base}}-P_{\text{range}}, P_{\text{base}}+P_{\text{range}}]$ with the constant $P_{\text{range}}$ defining the size of the range. We map the agent's actions $\textbf{a}_p$ from the range $[-1, 1]$ to each parameter's range. $P_{\text{base}}$ and $P_{\text{range}}$ are hyperparameters of both controllers.

\image{}{80}{Proposed scheme for learning PID parallel position/force control. The RL agent controls the controller parameters PD gains, PI gains, and the selection matrix, $S$.}

\subsubsection{Admittance Control}\label{subsec:admittance-scheme}
    is used to achieve a desired dynamic interaction between the manipulator and its environment.
    The admittance controller for position-controlled robots implemented is based on \cite{impedace1998implementation}. The admittance control is implemented on task-space instead of the robot joint-space. It follows the conventional control law
    \begin{equation}
        F_{ext} = m_d\Ddot{x}+b_d\Dot{x}+k_dx
    \end{equation}
    
    where $m_d$, $b_d$, and $k_d$ represent the desired inertia, damping, and stiffness matrices respectively. $F_{ext}$ is the actual contact force vector. $x$, $\Dot{x}$, $\Ddot{x}$ are the displacement of the manipulator's end-effector, its velocity and acceleration respectively.
    
    The admittance relationship can be expressed in Laplace-domain, adopting conventional expression of a second-order system as
    \begin{equation}
    \begin{aligned}
        \frac{X}{F}(s) & ~=  \frac{1/m_d}{s^2 + 2\zeta \omega_ns + \omega_n}
    \end{aligned}
    \end{equation}

    where $\zeta$ is the damping ratio and $\omega_n$ is the natural frequency, and they can be expressed by the admittance parameters as
     \begin{equation}\label{eq:damping-ratio}
      \begin{split}
            \zeta = \frac{b_d}{2\sqrt{k_d\,m_d}}
      \end{split}
      \quad \quad
      \begin{split}
            \omega_n = \sqrt{\frac{k_d}{m_d}}
      \end{split}
    \end{equation}
    We are proposing a variable admittance controller, where the inertia, damping, and stiffness parameters are learned by the RL agent. Additionally, a PD controller is included in our admittance control. The PD controller with the policy action, $\textbf{a}_x$, generates the nominal trajectory as explain in \Cref{subsec:system-overview}. The complete admittance control scheme is depicted in \Cref{fig:admittance-scheme}. The PD gains are also controlled by the policy at each time step.
\image{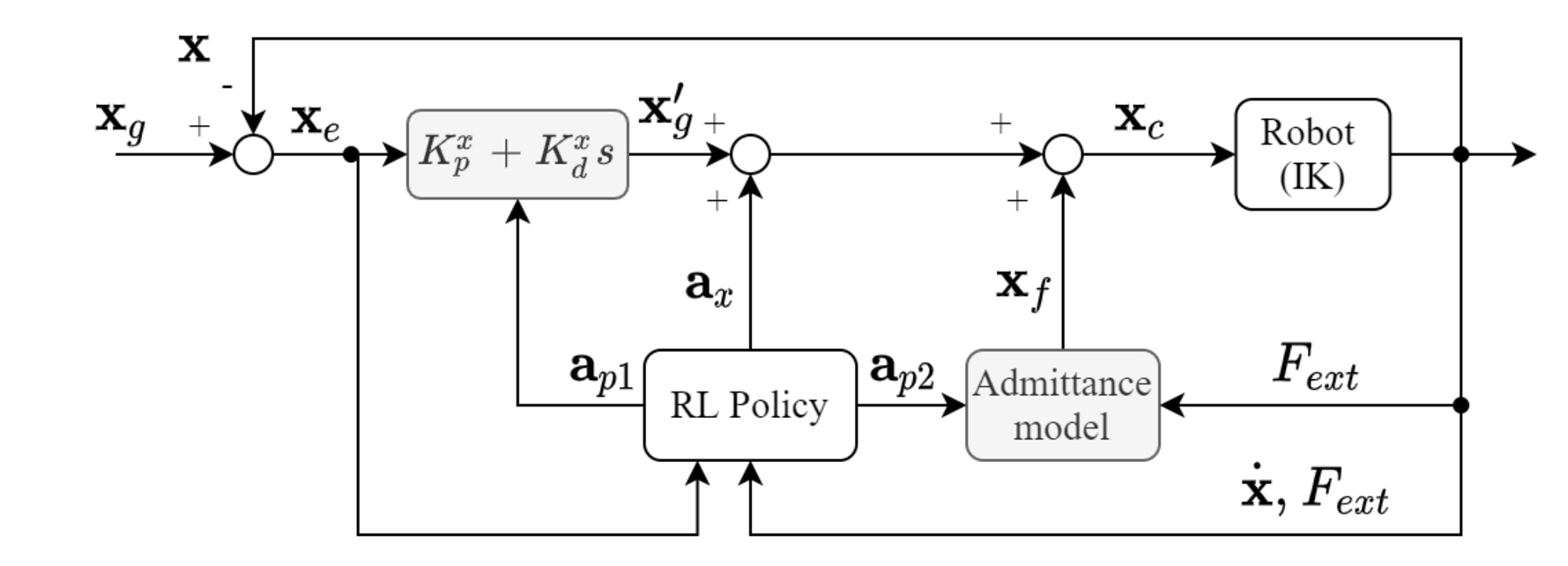}{80}{Proposed scheme for learning admittance control. A PD controller is included to regulate the input reference motion trajectory. The RL agent controls the PD gains, as well as, the admittance model parameters (inertia, damping and stiffness).}
    
    For the admittance control scheme, there are a total of 30 parameters; 12 from the position PD controller's gains and 18 from the inertia, damping, and stiffness parameters. Similarly, as mentioned in \Cref{subsec:parallel-scheme}, we reduced the number of controllable parameters to prevent unstable behavior of the robot and reduce the system's complexity. Following the same strategy described in \Cref{subsec:parallel-scheme}, of the PD controller, only the proportional gain, $K_p^x$, is controllable. Additionally, we considered the inertia parameter for each direction as a constant, $0.1$ kg$\cdot$m$^2$ in all our experiments as a similar payload is used across tasks. Furthermore, we compute the damping with respect to the inertia parameter and the stiffness parameter by defining a constant damping ratio. From \eqref{eq:damping-ratio} we have that

    \[
        b_d = 2\, \zeta \sqrt{k_d*m_d}
    \]

    Therefore, only the stiffness parameters are controllable. In total, the controllable parameters of the admittance control are reduced to 12 parameters; 6 PD gains and 6 stiffness parameters.

    In summary, the policy actions regarding the admittance controller's parameters are
    $\textbf{a}_p = [K_p^x, k_d]$

\subsection{Fail-safe mechanism}\label{subsec:safe-framework}

    Most modern robot manipulators already include a layer of safety in the form of an emergency stop. Nonetheless, the emergency stop exists at the extreme ends of the robot limits and completely interrupts the entire training session if triggered. To reactivate the robot, a human operator is required. To alleviate this inconvenience, we propose a mechanism that allows the robot to operate within less extreme limits. Thus, training of an RL agent can be done directly on the position-controlled manipulator with minimal human supervision.
    
    Our system controls the robot as if teleoperating it by providing a real-time stream of task-space motion commands for the robot to follow. Therefore, we added our safety layer between the streamed motion command and the robot's actual actuation. The fail-safe mechanism validates that the intended action is within a defined set of safety constraints. As shown in \Cref{alg:safe-manipulation}, for each action we check whether an IK solution exists for the desired position command, $\textbf{x}_c$, if so, whether the joint velocity required to achieve the IK solution, $\textbf{q}_c$, is within the speed limit. 

    If any of these validations are not satisfied, the intended action is not executed on the robot, and the robot remains in its current state for the present time step.
    Finally, we check if the contact force at the robot's end-effector is within a defined range limit. If not, the episode ends immediately. 

    The first two validations are proactive and prevent unstable behaviors of the manipulator before they occur. In contrast, the third validation is reactive, i.e., only after a collision has occurred (the force limit has been violated), the robot is prevented from further actions.

\begin{algorithm}[h]
\caption{Safe Manipulation Learning}\label{euclid}
\begin{algorithmic}[1]
\State \textit{Define} joint velocity limit $\Dot{\textbf{q}}_{max}$
\State \textit{Define} contact force limit $F_{max}$
\State \textit{Define} initial state $\textbf{x}_{0}$
\State \textit{Define} goal state $\textbf{x}_g$
\For{$n=0,\dotsi, N-1$ episodes}
    \For{$t=0,\dotsi, T-1$ steps}

            \State \textit{Get} current contact force: $F_{ext}$
            \State $\textbf{x}_e = \textbf{x}_g-\textbf{x}$
            \State \textit{Get} Observation: $\textbf{o} = [\textbf{x}_e, \Dot{\textbf{x}}, F_{ext}]$
            \State \textit{Compute} policy actions: $\pi_{\theta}(\textbf{a}_x, \textbf{a}_p|\textbf{o})$
            \State $\textbf{x}_c =$ \textit{control\_method}$(\textbf{x}_e, \textbf{a}_x,  \textbf{a}_p, F_{ext})$
            \State $\textbf{q}_c = $ \textit{IK\_solver}$(\textbf{x}_c)$
            \If {$\textbf{q}_c$ not exists} \textbf{continue}
            \EndIf
            \If {$|(\textbf{q}_t - \textbf{q}_c)/dt| > \Dot{\textbf{q}}_{max}$ } \textbf{continue}
            \EndIf
            \If {$F_{ext} > F_{max}$} \textbf{break}
            \EndIf

            \State \textit{Actuate} $\textbf{q}_c$ on robot
    \EndFor
    \State Reset to $\textbf{x}_0$
\EndFor
\end{algorithmic}
\label{alg:safe-manipulation}
\end{algorithm}
\squeezeup 
\subsection{Task's reward function}\label{subsubsec:cost-function}
For all the manipulation tasks considered, the same reward function was used:
\begin{equation}
\begin{aligned}
    {r}(\textbf{s},\textbf{a}) = & w_{1}L_m(\|{\textbf{x}_{e}/\textbf{x}_{max}}\|_{1,2}) + w_{2}L_m(\|\textbf{a}/\textbf{a}_{max}\|_2) +\\
                    & w_{3}L_m(\|F_{ext}/F_{max}\|_2) + w_4\rho + w_5\kappa
\end{aligned}
\label{eq:reward-function}
\end{equation}
where $\textbf{x}_{max}$, $\textbf{a}_{max}$, and $F_{max}$ are defined maximum values. $L_m(y) = y \mapsto x, x \in [1,0]$ is a linear mapping to the range 1 to 0, thus, the closer to the goal and the lower the contact force, the higher the reward obtained. $||\cdot||_{1,2}$ is L1,2 norm based on \cite{levine2016end}. The  $\textbf{x}_e$ is the distance between the manipulator's end-effector and the target goal at time step $t$. $\textbf{a}$ is the action taken by the agent. $F_{ext}$ is the contact force. $\rho$ is a penalty given at each time step to encourage a fast completion of the task. $\kappa$ is a reward defined as follows 
\begin{equation}
\kappa = \left\{\begin{matrix}
 200, & \textrm{Task completed}\\ 
 -10,     & \textrm{Safety violation} \\
   0,     & \textrm{Otherwise}
\end{matrix}\right.
\label{eq:safety-reward}
\end{equation}
Finally, each component is weighted via $w$, all $w$'s are hyperparameters. 
\color{black}
\section{Experiments}\label{sec:experiments}

    We propose a framework for safely learning manipulation tasks with position-controlled manipulators using RL. Two control schemes were implemented. With the following experiments, we seek to answer the following questions: Can a high-dimensional force controller be learned by the agent? Which action space, based on the number of adjustable controller's parameters provides the best learning performance?
    
    A description of the materials used for the experiments is given in \Cref{subsec:technical-details}. An insertion task was used for evaluating the learning performance of the RL agents with the proposed method on a simulated environment, described in \Cref{subsec:action-spaces}. Finally, the proposed method is validated on a real robot manipulator with high-precision assembly tasks. 

\subsection{Technical details}\label{subsec:technical-details}
    Experimental validation was performed both in a simulated environment using the Gazebo simulator \cite{koenig2004design} version 9 and on real hardware using the Universal Robot 3 e-series, with a control frequency of up to 500 Hz. The robotic arm has a Force/Torque sensor mounted at its end-effector and a Robotiq Hand-e gripper. Training was performed on a computer with CPU Intel i9-9900k, GPU Nvidia RTX-2800 Super.

\subsection{Action spaces for learning force control}\label{subsec:action-spaces}
    Each control scheme proposed in \Cref{sec:method} has a number of controllable parameters. The curse of dimensionality is a well known problem in RL \cite{sutton2018reinforcement}. Controlling few dimensions, number of parameters, makes the task easier to learn at the cost of losing dexterity. 
    
    In the following experiment, several policy models were evaluated. Each model has a different action space, i.e., a different number of controllable parameters. 
    We evaluate the learning performance of the models described in \Cref{table:action-spaces}, four models per control scheme. Each policy model has the same six parameters to control the position and orientation of the manipulator, $\textbf{a}_x$, but a different number of parameters to tune the controller's gains, $\textbf{a}_p$. From now on, we refer to each model by the name given in \Cref{table:action-spaces}. 

\begin{table}[b]
\caption{Policy models with different action spaces.}
\centering
\begin{tabular}{|c|c|c|c|c|c|}
\hline
\multirow{3}{*}{\textbf{\begin{tabular}[c]{@{}c@{}} \\ Control\\ Scheme\end{tabular}}} &
  \multirow{3}{*}{\textbf{Name}} &
  \multirow{2}{*}{\textbf{Pose}} &
  \multicolumn{3}{c|}{\textbf{Gains}} \\ \cline{4-6} 
 &
   &
   &
  \textbf{\begin{tabular}[c]{@{}c@{}}PD\end{tabular}} &
  \textbf{\begin{tabular}[c]{@{}c@{}}PI /\\ Stiffness\end{tabular}} &
  \textbf{\begin{tabular}[c]{@{}c@{}}Selection \\ Matrix S\end{tabular}} \\ \cline{3-6} 
                                    &        & $\textbf{a}_x$ & \multicolumn{3}{c|}{$\textbf{a}_p$} \\ \hline
\multirow{4}{*}{\textbf{Parallel}}    & P-9    & 6  & 1      & 1      & 1     \\ \cline{2-6} 
                                    & P-14   & 6  & 1      & 1      & 6     \\ \cline{2-6} 
                                    & P-19   & 6  & 6      & 6      & 1     \\ \cline{2-6} 
                                    & P-24   & 6  & 6      & 6      & 6     \\ \hline
\multirow{4}{*}{\textbf{Admittance}} & A-8    & 6  & 1      & 1      & -     \\ \cline{2-6} 
                                    & A-13   & 6  & 1      & 6      & -     \\ \cline{2-6} 
                                    & A-13pd & 6  & 6      & 1      & -     \\ \cline{2-6} 
                                    & A-18   & 6  & 6      & 6      & -     \\ \hline
\end{tabular}
\label{table:action-spaces}
\end{table}

    For a fair comparison, the action spaces were evaluated on a simulated peg-insertion environment so that we could guarantee the exact same initial conditions for each training session. The task is to insert a cube-shaped peg into a task board, where the hole has a clearance of $1$ mm. 
    Each policy model was trained for $50.000$ (50k) steps with a maximum of 150 steps per episode. The complete training session was repeated three times per model. Since the policy control frequency was set at $20$ Hz, each episode lasts a maximum of 7.5 seconds. The episode ends if 1) the maximum number of time steps is reached, 2) a minimum distance error from the target pose is achieved, 3) or if a collision occurs. In general, a complete training session takes about 50 minutes, including reset times.

\subsubsection*{\textbf{Results}}
    The comparison of learning curves for each policy model evaluated is shown in \Cref{fig:sim-reward-penalization}. In the figure, the average cumulative reward per episode across the training sessions (bold line) is displayed along with the standard deviation error (shaded colored area).  The results have been smoothed out using the exponential moving averages, with a 0.6 weight, to show the tendency of the learning curves.
    
\image{}{80}{Learning curve of training session with active penalization of violation of the safety constraints. Peg-insertion scenario on simulation.}
    From \Cref{fig:sim-reward-penalization}, the overall best performance is achieved with the policy models combined with the parallel control scheme. By the end of the training session, these families of policies can yield higher rewards than the policy models combined with the admittance control scheme. 
    
    For the parallel control scheme, the model with the worst performance is P-9; it can be seen that there is not enough control of the controller's parameters to learn a good policy consistently. On the other hand, the model P-24 has the slowest learning rate, but by the end of the training session, it can consistently learn a good policy. The policy model P-14 has the fastest learning rate and overall best performance. 
    
    For the admittance control scheme, the models A-13pd and A-18 have the best overall performance, with A-13pd yielding a cumulative reward as high as P-14 by the end of the training session. The model A-8, similar to P-9, has one of the worst performance; again, the lack of controllable parameters seems to have a big impact on learning a successful policy.

    It is worth noting that for both control schemes, the models P-14 and A-13pd have the best overall performance. They provide the best trade-off between system complexity and learn-ability. On the other hand, the models with the largest number of parameters P-24 and A-18 can learn successful policies, but they require a longer training time to achieve it.

    The parallel models' learning curve has larger standard deviation. One factor that contributes to these results is the selection matrix $S$, which highly affects the performance of the controller. Small changes of this parameter can make the behavior completely different. The agent's random exploration of this parameter can result in very different results during the learning phase.

\subsection{Safe learning}
    The developed fail-safe mechanism was not only evaluated as a mechanical safety that enables the real robot to explore random action without human supervision. We validate the usefulness of providing information to the robot about the safety constraints violations. Thus, we compare the proposed reward function \Cref{eq:reward-function} with a variant that does not provide any punishment when a safety constraint is violated, i.e., $\kappa$ gives a reward if the task is completed or zero otherwise, see \Cref{eq:safety-reward}. We trained all policy models with this modified reward function.

\image{}{80}{Learning curve of training without penalizing violation of safety constraints on the reward function. Peg-insertion scenario on simulation.}
\begin{table}
\centering
\caption{Collision detected during training session.}
\begin{tabular}{|c|c|c|c|} 
\hline
\multirow{2}{*}{\textbf{Model}} & \multicolumn{3}{c|}{\textbf{avg. \# of collisions across training sessions}} \\ 
\hhline{|~---|}
 & \textbf{Penalization} & \textbf{No penalization} & \textbf{Difference} \\ 
\hhline{|----|}
A-8 & 326 & 455 & -39\% \\ 
\hline
A-13 & 350 & 408 & -16\% \\ 
\hhline{|----|}
\textbf{A-13pd} & \textbf{300} & \textbf{462} & \textbf{-54}\% \\ 
\hline
A-18 & 451 & 457 & -1\% \\ 
\hline
P-9 & 187 & 369 & -98\% \\ 
\hline
\textbf{P-14} & \textbf{121} & \textbf{206} & \textbf{-70}\% \\ 
\hline
P-19 & 183 & 392 & -115\% \\ 
\hline
P-24 & 219 & 337 & -43\% \\
\hline
\end{tabular}
\label{table:collisions}
\end{table}
\subsubsection*{\textbf{Results}}
    \Cref{fig:sim-reward-no-penalization} shows the comparison of the learning curves of all models with a reward function that does not penalize violation of safety constraints. The results clearly show that the overall performance considerably decreases. The learning speed also decreases, as can be noted by comparing the performance of, for example, the model A-13pd. Learning with active penalization helps the agent learn policies that yield rewards of +100 by 12,000 steps while it takes as much as 20,000 steps without penalization to achieve similar performance. Parallel control models show similar results. Moreover, the learning curves are nosier, meaning that the models can not reliably find a successful policy. 
    
    Additionally, we counted the average number of collisions detected during training sessions for each policy model. \Cref{table:collisions} shows the training session results using the proposed reward function with active penalization of the safety constraints and the reward function without. In all cases, we see a high decrease in the number of collisions when collisions are penalized. In other words, the training session can be considered safer when the robot gets feedback on the undesired outcomes, when safety constraints are violated. Particularly, in the case of the parallel control scheme, the models have difficulty understanding that collisions are a poor behavior; thus, those models keep getting stuck on episodes that finish too soon due to collision. Moreover, the models A-13pd and P-14 do not only learn faster than other models but also produce the lowest number of collisions within their family of policies. On the other hand, the policy models with the highest number of parameters, A-18 and P-24, are able to learn successful policies at the cost of producing the highest number of collisions.

\image{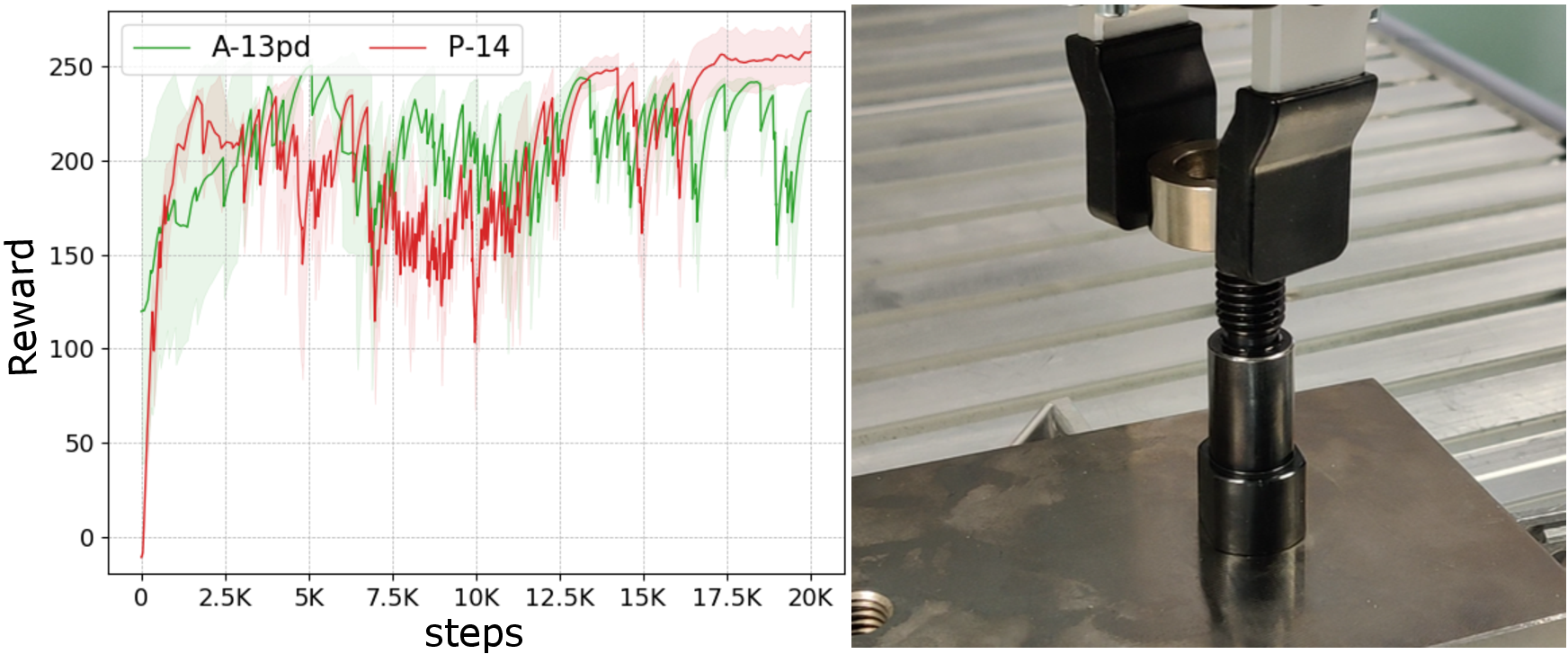}{80}{Ring-insertion task. Hole clearance of 0.2 mm. Cumulative reward per step of 20,000-steps training sessions of A-13pd and P-14 policy models.}  

\subsection{Real robot experiments}\label{subsec:real-robot-experiments}

    Our proposed method was validated on real hardware using two high-precision assembly tasks. The first task involves an insertion task of a metallic ring into a bolt with a clearance of 0.2 mm, as shown in \Cref{fig:real-ring2}. The second task is a more precise insertion task of the metallic peg into a pulley, with a clearance of 0.05 mm, as shown in \Cref{fig:real-wrs2}. Another robotic arm holds the pulley, and the center of the pulley is slightly flexible, which makes contact less stiff than the ring-insertion task. However, since the clearance is smaller, the peg is likely to get stuck if the peg is not adequately aligned, increasing the difficulty of solving the task. The best policy models from the previous experiment were used for training, P-14, and A-13pd. Both models were trained for 20,000 steps, twice. The episodes have a maximum length of 200 steps, about 10s.

\subsubsection{\textbf{Ring-insertion task results}} 
    From \Cref{fig:real-ring2}, both models A-13pd and P-14 can quickly learn successful policies. The high stiffness of the ring and bold makes the task more likely to result in a collision. The model P-14 produced an average of 45 collisions per training session, while A-13pd produced 34. Despite firmly grasping the ring with the robotic gripper, the position/orientation of the ring can still slightly change. These slight changes can explain the drops in performance during the training session. However, the agents can adapt and learn to succeed in the task.

\subsubsection{\textbf{Peg-insertion task results}}
    From \Cref{fig:real-wrs2}, we can see that it takes a lot more learning time to find a successful policy for both policy models compare to the ring-insertion task. While both policy models find a successful policy after about 13k steps, A-13pd achieved better consistent performance. As mentioned above, the physical interaction for this task is less stiff; thus, the average collisions per training session were fewer than in the ring-insertion task. For models A-13pd and P-14, the average number of collisions was 4 and 26, respectively.
\image{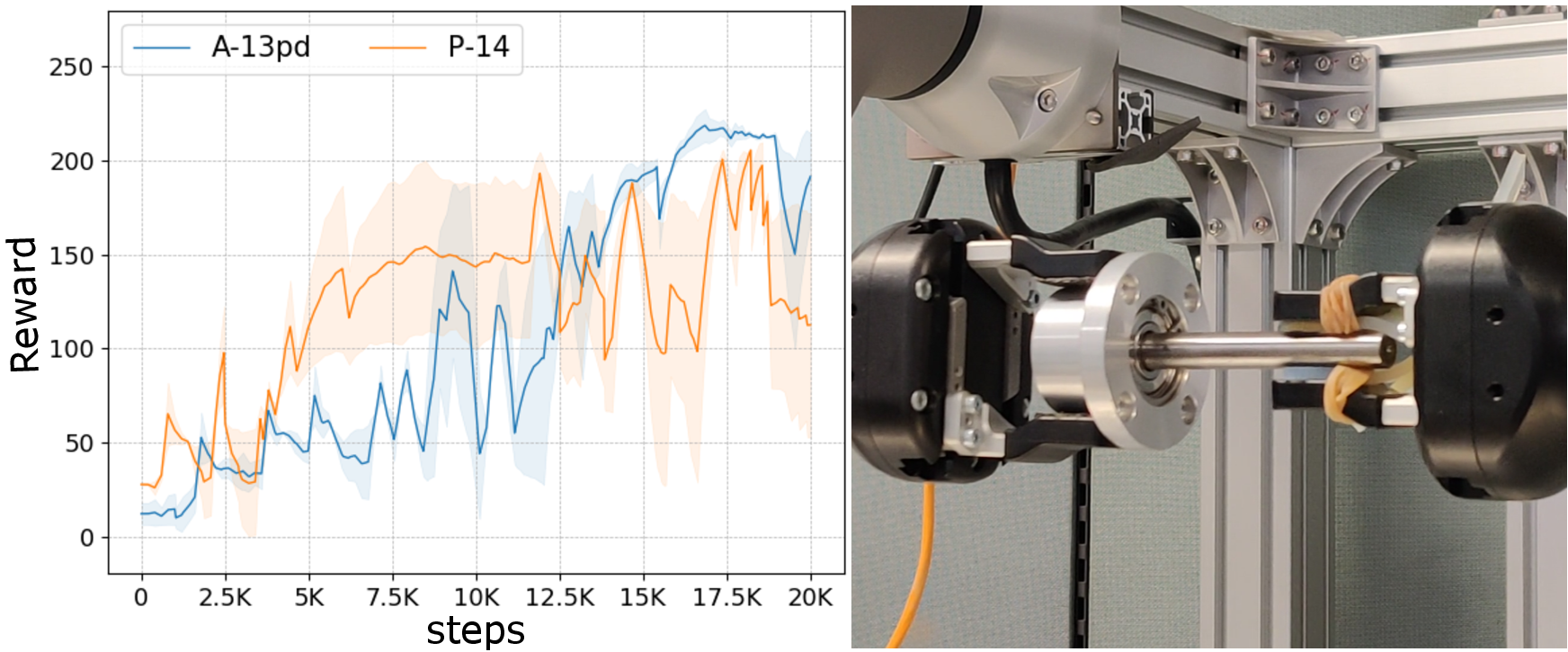}{80}{Peg-insertion task. Hole clearance of 0.05 mm. Cumulative reward per step of 20,000-steps training sessions of A-13pd and P-14 policy models.}
    
    The evolution of the policy model A-13pd, across a training session, is shown in \Cref{fig:wrs-imp}. The figure displays the observation per time step of only the insertion direction. The actions, $\textbf{a}_x$ and $\textbf{a}_p=[K_p^x, k_d]$ are also displayed. Observations and actions have been mapped to a range of [1, -1]. The peg-insertion task has three phases. A search phase before contact (Yellow). A search phase after initial contact (Red). An insertion phase (Green). On the left, the initial policy, we can clearly see that the insertion was not successful even after 200 steps, as well as a rather random selection of actions. On the contrary, on the right side, the task is being solved at around 130 steps. On top of that, the controller's parameters $k_d$ and $K_p^x$ have a clear response to the contact force perceived. After the first contact with the surface (Red), $k_d$ and $K_p^x$ are dramatically reduced, as a result, decreasing motion speed and reducing stiffness of the manipulator, which reduces the contact force. Then, when the peg is properly aligned (Green), $k_d$ and $K_p^x$ are increased to apply force to insert the peg -against the friction of the insertion- and to finish the task faster.

\begin{figure}
    \centering
    \includegraphics[width=\columnwidth,height=40mm]{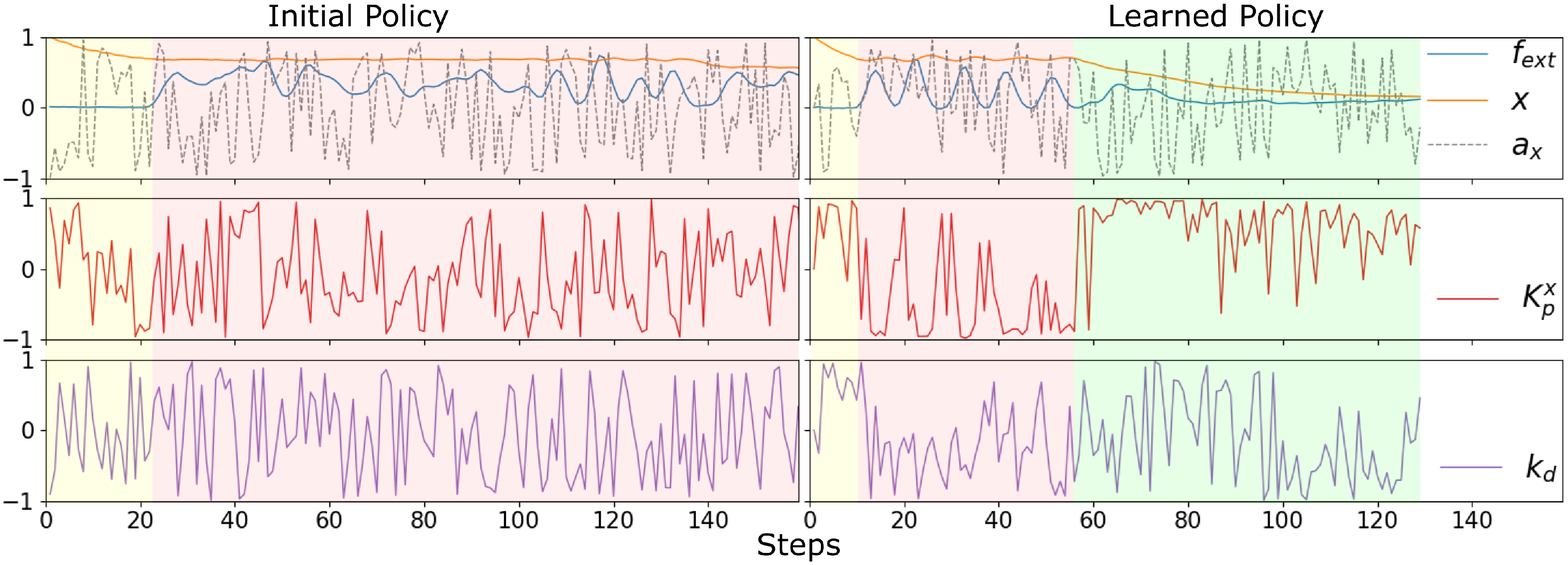}
    \caption{\textbf{A-13pd}: policy performance evolution on peg-insertion task. On the left, performance of the initial policy tried by agent. On the right, performance of the learned policy after training. All values correspond to the insertion direction only. Only 160 steps are displayed for space constraints. Insertion task divided into three phases: a search phase before contact (Yellow), a search phase after initial contact (Red) and an insertion phase (Green).}
    \label{fig:wrs-imp}
\end{figure}

\section{Discussion}\label{sec:discussion}

    In this work, we have presented a framework for safely learning contact-rich manipulation tasks using reinforcement learning with a position-controlled robot manipulator. The agent learns a control policy that defines the motion trajectory, as well as fine-tuning the force control parameters of the manipulator's controller. We proposed two learning force control schemes based on two standard force control methods, parallel position/force control, and admittance control. To validate the effectiveness of our framework, we performed experiments in simulation and with a real robot.

    First, we empirically study the trade-off between control complexity and learning performance by validating several policy models, each with a different action space, represented by a different number of adjustable force control parameters. Results show that the agent can learn optimal policies with all policy models considered, but the best results are achieved with the models A-13pd and P-14. These models yield the highest reward during training, proving to be the best trade-off between system complexity and learn-ability.
    
    Second, results on a real robot showed the effectiveness of our method to safely learn high-precision assembly tasks on position-controlled robots. The first advantage is that the fail-safe mechanism allows for training with minimal human supervision. The second advantage is that including information about the violation of safety constraints on the reward function helps speed up learning and reduce the overall number of collisions occurred during training.
    
    Finally, in the peg insertion task, the motion trajectory is essential when the robot is in the air, while the force control parameters become essential when the peg is in contact with a surface or the hole. Results show that our framework can learn policies that behave accordingly on the different phases of the task. The learned policies can simultaneously define the motion trajectory and fine-tune the compliant controller to succeed in high-precision insertion tasks.
    
    One of the limitations of our proposed method is that the performance is highly dependent on the choice of the controller's hyperparameters, more specifically, the base and range values of the controller's gains. In our experiments, we empirically defined said hyperparameters. However, to address said limitation, an interesting avenue for future research is to obtain these hyperparameters from human demonstrations, and then refine the force control parameters using RL. Additionally, for simplicity, we assume knowledge of the goal pose of the end-effector for each task. However, vision could be used to get a rough estimation of the target pose to perform an end-to-end learning, from vision to low-level control, as proven in previous work \cite{levine2016end}.
\squeezeup


\bibliographystyle{IEEEtran}
\bibliography{main}




\end{document}